\documentclass[10pt,twocolumn,letterpaper]{article}
\usepackage[dvipsnames,table]{xcolor}
\usepackage[pagenumbers]{iccv} 
\usepackage{booktabs}
\usepackage{multirow}
\definecolor{LightBlue}{rgb}{0.9,0.94,1}
\usepackage{fp}
\usepackage{adjustbox}
\usepackage{pifont}

\definecolor{iccvblue}{rgb}{0.21,0.49,0.74}
\usepackage[pagebackref,breaklinks,colorlinks,allcolors=iccvblue]{hyperref}

\def\paperID{2714} 
\def\confName{ICCV}
\def\confYear{2025}

\title{Insert Anything: Image Insertion via In-Context Editing in DiT}

\author{
    Wensong Song$^{1}$\quad
    Hong Jiang$^{1}$\quad
    Zongxing Yang$^{2}$\quad
    Ruijie Quan$^{3}$\quad
    Yi Yang$^{1}$\quad\\
    $^{1}$Zhejiang University \quad
    $^{2}$Harvard University \quad
    $^{3}$Nanyang Technological University \\
    \url{https://song-wensong.github.io/insert-anything/}
}

\begin{document}

\twocolumn[{
\renewcommand\twocolumn[1][]{#1}
\maketitle
\begin{center}
\centering
  \centering
  \vspace{-2em}
  \includegraphics[width=1\linewidth]{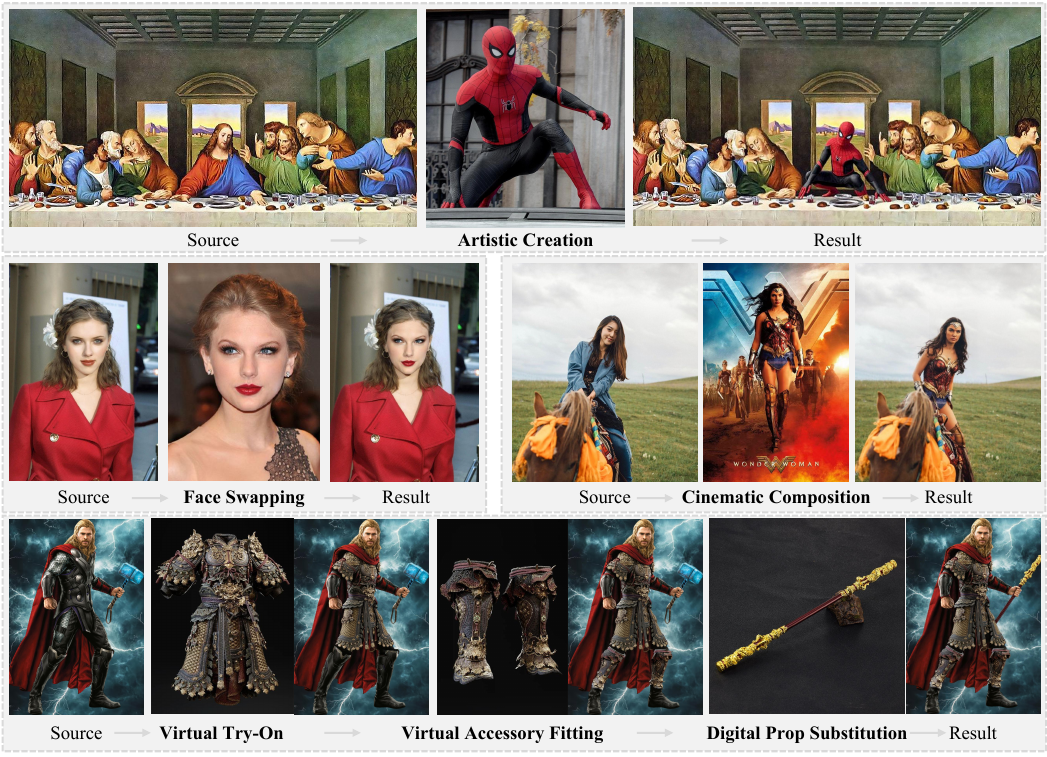}
  \captionsetup{skip=1pt}
  \vspace{-1em}
  \captionof{figure}{\textbf{Applications of Insert Anything.} Our unified image insertion framework supports diverse practical scenarios, including artistic creation, realistic face swapping, cinematic scene composition, virtual garment try-on, accessory customization, and digital prop replacement, demonstrating its versatility and effectiveness in various image editing tasks.}
  \label{fig:teaser}
\end{center}
}]

\begin{abstract}
This work presents \textbf{Insert Anything}, a unified framework for reference-based image insertion that seamlessly integrates objects from reference images into target scenes under flexible, user-specified control guidance. Instead of training separate models for individual tasks, our approach is trained once on our new \textbf{AnyInsertion} dataset--comprising 120K prompt-image pairs covering diverse tasks such as person, object, and garment insertion--and effortlessly generalizes to a wide range of insertion scenarios. Such a challenging setting requires capturing both identity features and fine-grained details, while allowing versatile local adaptations in style, color, and texture. To this end, we propose to leverage the multimodal attention of the Diffusion Transformer (DiT) to support both mask- and text-guided editing. Furthermore, we introduce an in-context editing mechanism that treats the reference image as contextual information, employing two prompting strategies to harmonize the inserted elements with the target scene while faithfully preserving their distinctive features. Extensive experiments on AnyInsertion, DreamBooth, and VTON-HD benchmarks demonstrate that our method consistently outperforms existing alternatives, underscoring its great potential in real-world applications such as creative content generation, virtual try-on, and scene composition.
\end{abstract}    
\section{Introduction}
\label{sec:intro}
Recent advances in diffusion models~\cite{ho2020denoising,peebles2023scalable} have revolutionized image editing~\cite{brooks2023instructpix2pix, zhang2023magicbrush, wang2024taming, shen2024audioscenic, li2024drip, xu2024gg}, enabling tasks such as style transfer~\cite{chung2024style, zhang2023inversion} and inpainting~\cite{xie2023smartbrush, lugmayr2022repaint}.
Among various editing techniques, reference image-based editing plays a pivotal role by providing explicit visual cues that guide the editing process.
In this approach--also referred to as image insertion, a target image is modified by seamlessly incorporating elements from a reference image, ensuring that the overall coherence and quality remain intact.
Unlike text-based instructions, reference images offer concrete details regarding style, color, and texture, making them indispensable for achieving contextually consistent results.

Despite promising advances that have been made, several challenges remain in current image insertion work~\cite{chen2024anydoor, chen2024zero, mao2025ace++, parihar2024text2place, chong2024catvton,kulal2023putting, xu2024ootdiffusion,he2024freeedit}. 
\ding{182} \textbf{Task-Specific Focus.} 
Both approaches and datasets of them aim to address only specific tasks, such as either person insertion~\cite{parihar2024text2place, kulal2023putting} or garment insertion~\cite{chong2024catvton, xu2024ootdiffusion}, which limits their broader applicability to real-world scenarios.
\ding{183} \textbf{Fixed Control Mode.} These methods typically rely on inflexible control mode, \eg, supporting either mask-guided editing~\cite{chong2024catvton,xu2024ootdiffusion,chen2024anydoor,chen2024zero,kulal2023putting} that uses manually provided masks to delineate the editing region,
or text-guided editing~\cite{shen2025tarpro,he2024freeedit,mao2025ace++}
that depends on the language instructions to control editing. This rigidity in control modes hinders creative flexibility.
\ding{184} \textbf{Inconsistent Visual-Reference Harmony.} Even when these methods~\cite{chen2024anydoor, he2024freeedit,chen2024zero, mao2025ace++} succeed in inserting new elements, they often struggle to maintain visual harmony between the inserted content and the target image, while ensuring that the distinctive characteristics of the reference are retained. As a result, outputs usually exhibit artifacts or stylistic mismatches that compromise overall quality and authenticity.

First, to address the challenge \ding{182}, we introduce \textit{\textbf{AnyInsertion}}, a large-scale dataset, specifically designed for reference image-based image editing. \textit{\textbf{AnyInsertion}} is characterized by two key features:
1) It supports a wide range of image insertion tasks, including person insertion, object insertion, and garment insertion, by providing diverse categories such as human subjects, daily necessities, garments, furniture, and other objects. This diversity ensures broader applicability to real-world scenarios.
2) It contains 159k prompt-image pairs, consisting of 58k mask-prompt pairs and 101k text-prompt pairs. These diverse control modes enable the dataset to provide flexible training and evaluation for both mask-guided and text-guided image editing tasks.

Building on our dataset, we introduce \textbf{Insert Anything}, a unified framework that is capable of inserting \textit{any} object from a reference image into a target scene while supporting \textit{multiple control modes} (\ie, mask and text).
To address the challenge \ding{183}, we propose to leverage the multimodal attention of the Diffusion Transformer (DiT)~\cite{peebles2023scalable} to jointly model the relationships among text, masks, and image patches, enabling flexible control over image editing tasks guided by either masks or text.
Moreover, to address the challenge \ding{184}, we introduce \textit{in-context editing}, a new approach that treats the reference image as contextual content rather than as a standalone input.
It allows for interactions between the inserted elements and their surrounding context, enabling the model to capture inherent correlations implicitly. As a result, it ensures faithful preservation of identity features and seamless integration of the inserted elements into the target scene.
For the two editing modes, mask-prompt editing fills the target image’s masked regions with elements from the reference image, while text-prompt editing modifies the source image based on a text description of the reference. 
To accommodate these distinct control modes, we design two prompting strategies:
\texttt{mask-prompt diptych} for mask-guided insertion and \texttt{text-prompt triptych} for text-guided insertion.
In \texttt{mask-prompt diptych}, the left panel displays the reference image containing the desired elements, and the right panel shows the masked source image. In \texttt{text-prompt triptych}, the left panel shows the reference image, the middle one presents the source image, and the right one contains the inpainted result generated from text descriptions.

We perform extensive evaluations on our proposed \textit{\textbf{AnyInsertion}} dataset as well as on two additional benchmarks--DreamBooth~\cite{ruiz2023dreambooth}, and VTON-HD~\cite{choi2021viton}. Our experimental results on human, object, and garment insertion tasks across these datasets consistently demonstrate that our method achieves state-of-the-art performance.
In summary, our \textbf{contributions} are three-fold:
\begin{itemize}
    \item We introduce \textit{\textbf{AnyInsertion}}, a large-scale dataset containing 120K prompt-image pairs, spanning a wide range of insertion tasks, \ie person, object, and garment insertion.
    \item We propose \textbf{Insert Anything}, a unified framework that seamlessly handles multiple insertion tasks (person, object, and garment) through a single model.
    \item We are the first, to our best knowledge, to leverage the DiT for image insertion, exploiting its unique capabilities for different control modes. 
    \item We develop \textit{in-context editing}, employing \texttt{diptych} and \texttt{triptych prompting} to seamlessly integrate reference elements into target scenes while preserving identity.
\end{itemize}
\section{Related Work}
\label{sec:related_work}

\noindent \textbf{Image Insertion.} 
Image insertion methods are usually categorized by task specificity and control strategy. Task-specific approaches include person insertion and garment insertion.
In person insertion, Kulal~\etal~\cite{kulal2023putting} introduced an inpainting-based method, while ESP~\cite{ostrek2024synthesizing} generates personalized figures guided by 2D pose and scene context. Text2Place~\cite{parihar2024text2place} leverages SDS loss to optimize semantic masks for accurate human placement.
For garment insertion, OOTDiffusion~\cite{xu2024ootdiffusion} employs a ReferenceNet structure similar to a denoising UNet for processing garment images, whereas CatVTON~\cite{chong2024catvton} spatially concatenates garment and person images to enable lightweight virtual try-on.
Among general object editing methods~\cite{ chen2024anydoor, song2024imprint, chen2024zero, he2024freeedit}, AnyDoor~\cite{chen2024anydoor} and MimicBrush~\cite{chen2024zero} both support mask-guided insertion. AnyDoor utilizes DINOv2~\cite{oquab2023dinov2} for feature extraction and ControlNet~\cite{zhang2023adding} to preserve high-frequency details, whereas MimicBrush uses a UNet~\cite{ronneberger2015u} to extract reference features while maintaining scene context via depth maps and unmasked background latents. Freeedit~\cite{he2024freeedit} supports text-guided insertion through multi-modal instructions combined with UNet-based fine-detail extraction.
Our approach differs from these methods by leveraging in-context learning for efficient high-frequency detail extraction, eliminating the need for additional networks like ControlNet~\cite{zhang2023adding}, and supporting both mask and text prompts.

\noindent \textbf{Reference-Based Image Generation.}
Reference-based image generation, also known as subject-driven generation, takes as input a reference image containing the subject and aims to generate a new image that faithfully preserves the subject's details, such as facial features~\cite{wang2024instantid} (Wang et al., 2024) or stylistic attributes~\cite{sohn2023styledrop} (Sohn et al., 2023). These methods fall into two main categories: those requiring test-time fine-tuning (\eg, Textual Inversion and DreamBooth~\cite{ruiz2023dreambooth}) and those that adopt a training-based approach (e.g., IP-Adapter~\cite{ye2023ip}). Notable recent innovations include In-context LoRA~\cite{huang2024context}, which leverages DiT's~\cite{peebles2023scalable} in-context learning capabilities for thematic image generation, and Diptych Prompt~\cite{shin2024large}, which enables training-free zero-shot reference-based image generation through the Flux ControlNet inpainting model.
Unfortunately, these methods lack the capability for controllable image insertion; they can only generate the subject's context via text descriptions. To address this limitation, we propose Insert Anything, which extends DiT's in-context learning abilities to the image insertion editing scenario by providing the subject's context through images. Additionally, we introduce the \textit{AnyInsertion} dataset—the first open-source dataset for image insertion.

\noindent \textbf{Unified Image Generation and Editing.}
Recent frameworks have attempted to unify multiple image generation and editing tasks. ACE~\cite{han2024ace}, which employs a conditioning unit for multiple inputs; Qwen2vl-flux~\cite{erwold-2024-qwen2vl-flux} uses a vision-language model for unified prompt encoding; and OminiControl~\cite{tan2024ominicontrol} concatenates condition tokens with image tokens.
AnyEdit~\cite{yu2024anyedit}, Unireal~\cite{chen2024unireal}, and Ace++~\cite{mao2025ace++} provide partial support for image insertion tasks, but none offers a comprehensive solution for all three insertion types with both mask and text prompt support, which distinguishes our Insert Anything framework.
\begin{figure*}[htbp]
    \centering
    \captionsetup[subfigure]{skip=1pt, aboveskip=0pt,belowskip=0pt}
    \begin{subfigure}[t]{0.62\textwidth}
        \centering
        \includegraphics[width=\linewidth]{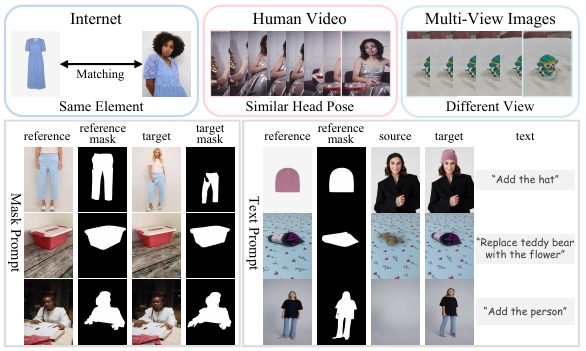}
        \caption{Data source and example. Image pairs are collected from internet sources, human videos, and multi-view images. The dataset is divided into mask-prompt and text-prompt categories, with further subdivisions into accessories, objects, and persons for each prompt type.}
        \label{fig:data-source-example}
    \end{subfigure}
    \hfill
    \begin{subfigure}[t]{0.36\textwidth}
        \centering
        \includegraphics[width=\linewidth]{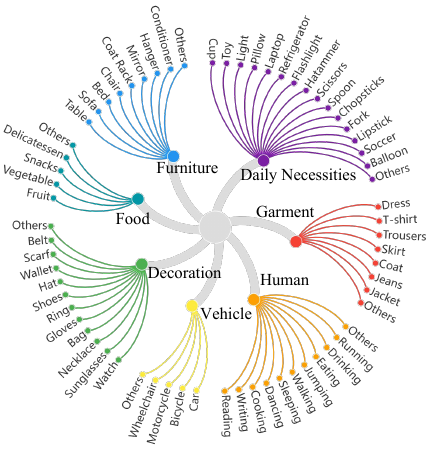}
        \caption{Visualization of dataset categories. Covers diverse insertion scenarios: furniture, daily necessities, garments, vehicles, and humans.}
        \label{fig:data-info}
    \end{subfigure}%
    \vspace{-2mm}
    \caption{Overview of AnyInsertion dataset, highlighting its source and example (a) and diversity (b).}
    \vspace{-10pt}
\end{figure*}

\section{AnyInsertion Dataset}
\label{sec:dataset}
To enable diverse image insertion tasks, we introduce a new large-scale dataset, AnyInsertion. 
Comparison with existing datasets is first described in \S\ref{sec:data_comparison}. The dataset construction process is then detailed in \S\ref{sec:data_construction}. Finally, detailed dataset statistics are provided in \S\ref{sec:dataset_info}.

\subsection{Comparison with Existing Datasets}
\label{sec:data_comparison}
Existing datasets suffer from several limitations:
\textbf{Limited Data Categories.} FreeEdit~\cite{he2024freeedit} dataset primarily focuses on animals and plants, and the VITON-HD~\cite{choi2021viton} dataset specializes in garments. Even AnyDoor~\cite{chen2024anydoor} and MimicBrush~\cite{chen2024zero} include a large scale of data, they contain only very few samples related to person insertion. \textbf{Restricted Prompt Types.} FreeEdit~\cite{he2024freeedit} provides only text-prompt data, while VITON-HD supports only mask-prompt data.
\textbf{Insufficient Image Quality.} AnyDoor and MimicBrush utilize a large volume of video data. These video datasets often suffer from low resolution and motion blur.
To address these issues, we have constructed an AnyInsertion dataset. 

As shown in Table~\ref{tab:dataset_comparision}, compared to existing datasets~\cite{he2024freeedit, choi2021viton}, AnyInsertion encompasses diverse categories, offers higher resolution images, supports both mask- and text-prompts, and includes a larger number of samples.

\begin{table}[t]
\small
\setlength\tabcolsep{1.5pt}
\captionsetup{skip=1pt}
\centering
\caption{\textbf{Comparison of existing image insertion datasets with our AnyInsertion dataset.} AnyInsertion addresses the limitations of existing datasets by covering diverse object categories, supporting both mask- and text-prompt, and providing higher-resolution images suitable for various practical insertion tasks.}
\label{tab:dataset_comparision}
\begin{tabular}{lcccc}
    \toprule
    Dataset & Theme & Resolution & Prompt & \#Edits\\
    \midrule
    FreeBench~\cite{he2024freeedit}  & Daily Object & 256 x 256 &  Text & 131,160\\
    VITON-HD~\cite{choi2021viton}  & Garment & 1024 x 768 &  Mask & 11,647\\
    AnyInsertion & Multifield & Mainly 1–2K & Mask / Text & 159,908\\
    \bottomrule
\end{tabular}
\vspace{-10pt}
\end{table}

\subsection{Data Construction}
\label{sec:data_construction}

\noindent \textbf{Data Collection.}
\label{sec:data_collection}
Image insertion requires paired data: a reference image containing the element to be inserted, and a target image where the insertion occurs.

As illustrated in Fig.~\ref{fig:data-source-example}, we employ image matching techniques~\cite{lindenberger2023lightglue} to create paired target and reference images and gather corresponding labels from internet sources, leveraging the abundance of images showing accessories and people wearing them. For object-related data, we select images from MVImgNet~\cite{yu2023mvimgnet}, which provides varying viewpoints of common objects, to serve as reference-target pairs. For person insertion, we apply head pose estimation~\cite{cobo2024representation} to select frames with similar head poses but varied body poses from the HumanVid dataset~\cite{wang2024humanvid}, which offers high-resolution video frames in real-world scenes. Frames with excessive motion blur are filtered out using blur detection~\cite{pech2000diatom}, resulting in high-quality person insertion data.

\noindent \textbf{Data Generation.}
\label{sec:data_generation}
In our framework, image insertion is designed to support two control modes: mask-prompt and text-prompt. Mask-prompt editing requires a mask to specify the insertion region in the target image, using elements from the reference image to fill the masked area of the target image. Text-prompt editing requires text to describe how the reference image's elements are inserted into the source image to form the target image. The data pairs for mask-prompt and text-prompt editing differ in structure.
As shown in Fig.~\ref{fig:data-source-example}, AnyInsertion provides data for both mask- and text-prompt control modes. 

\noindent \textbf{\textit{Mask-Prompt.}}
For mask-prompt editing, each data sample is represented as a tuple: (reference image, reference mask, target image, target mask). Specifically, we use Grounded-DINO~\cite{liu2024grounding} and Segment Anything (SAM)~\cite{kirillov2023segment} to generate reference and target masks from input images and labels.

\noindent \textbf{\textit{Text-Prompt.}}
For text-prompt editing, each data sample is represented as a tuple: (reference image, reference mask, target image, source image, text). The source image, text description, and reference mask are derived from the reference and target images, as follows:

\noindent \textit{Source Image Generation.}
Source images are generated by applying replacement or removal operations to the target images. For replacements, we design category-specific instruction templates (\eg, \textit{``replace \texttt{[reference]} with \texttt{[source]}''}) and use a text-based editing model to generate initial edits. To maintain image harmony, we employ FLUX.1 Fill [dev]~\cite{flux2024} that preserves unedited regions while allowing modifications within masked areas. For removals, we use the DesignEdit~\cite{jia2024designedit} model with the target mask to obtain the removal result.

\noindent \textit{Text Creation.} For replacement operations, we adapt instruction templates to reflect the desired transformations, such as \textit{``replace \texttt{[source]} with \texttt{[reference]}''} for the transition from source to target. For addition operations, we use the format \textit{``add \texttt{[label]}''} to describe the transition.

\noindent \textit{Reference Mask Extraction.}
We extract reference masks using the same method as in Mask-Prompt Editing.

More details on dataset construction can be found in \S\ref{sec:more_dataset} in the supplementary material.

\begin{figure*}[h]
  \centering
   \includegraphics[width=\textwidth]{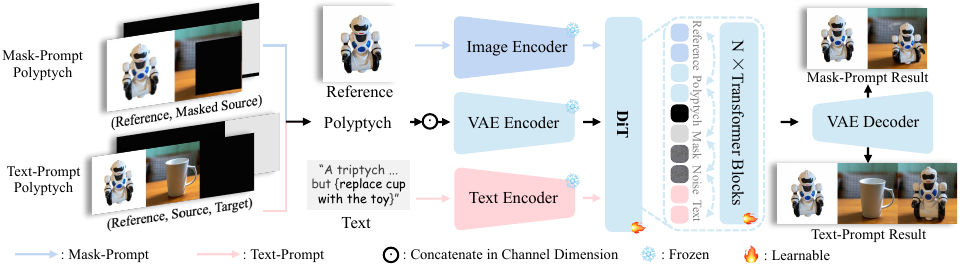}
   \captionsetup{skip=1.5pt}
   \caption{\textbf{Overview of the Insert Anything model framework.} Given different types of prompts, our unified framework processes polyptych inputs (concatenation of reference, source, and masks) through a frozen VAE encoder to preserve high-frequency details, and extracts semantic guidance from image and text encoders. These embeddings are combined and fed into learnable DiT transformer blocks for in-context learning, enabling precise and flexible image insertion guided by either mask- or text-prompt.}
   \label{fig:iam}
   \vspace{-10pt}
\end{figure*}

\subsection{Dataset Overview}
\label{sec:dataset_info}
AnyInsertion dataset consists of training and testing subsets. The training set includes 159,908 samples across two prompt types: 58,188 mask-prompt image pairs (reference images, reference masks, target images, and target masks) and 101,720 text-prompt image pairs (reference images, reference masks, source images, target images, and texts). As shown in Fig.~\ref{fig:data-info}, the dataset covers diverse categories including human subjects, daily necessities, garments, furniture, and various objects. This diversity enables the dataset to supports multiple insertion tasks including person insertion, object insertion, and garment insertion, thereby supporting a wide range of real-world applications. For evaluation, we curated a test set consisting of 158 data pairs: 120 mask-prompt pairs and 38 text-prompt pairs. The mask-prompt subset includes 40 pairs for object insertion, 30 pairs for garment insertion, and 60 pairs for person insertion (30 simple scene insertions and 30 complex scene insertions). The text-prompt subset contains 16 pairs for object insertion and 22 pairs for garment insertion.

\section{Insert Anything Model}
\label{sec:iam}

\textbf{Overview.} The image insertion task requires three key inputs: a reference image containing the element to be inserted, a source image providing the background context, and a control prompt (either mask or text) that guides the insertion process. The goal is to generate a target image that seamlessly integrates the element from the reference image (hereafter referred to as the “reference element”) into the source image while preserving the identity of reference element (\ie, the visual features that define the reference element), and adhering to the specifications in the prompt. As illustrated in Fig.~\ref{fig:iam}, our approach integrates three components: (1) a polyptych in-context format that organizes inputs to leverage contextual relationships (\S\ref{sec:method_in_context}), (2) semantic guidance mechanisms that extract high-level information from either text prompts or reference images, and (3) a DiT-based architecture that combines these elements through multimodal attention (\S\ref{sec:method_dit_inpainting}). Together, these components enable flexible control while maintaining visual harmony between inserted elements and their surrounding context.

\subsection{In-Context Editing}
\label{sec:method_in_context}
In-context editing involves integrating reference elements into a source image while maintaining the contextual relationships between them. To achieve this, we first perform a background removal step to isolate the reference element. Following the approach~\cite{chen2024anydoor, shin2024large}, we utilize the background removal process $R_{\text{seg}}$ using Grounding-DINO and SAM to remove the background of the reference image, leaving only the object to be inserted. 

Once the reference element is obtained, we perform in-context editing using two distinct approaches, corresponding to the mask-prompt and text-prompt modes.

\noindent \textbf{\texttt{Mask-Prompt Diptych}.}
For mask-prompted editing, we propose a two-panel structure (diptych), which concatenates the processed reference image with a partially masked source image:
\begin{equation}
    I_{\text{diptych}} = [R_{\text{seg}}(I_{\text{ref}}); I_{\text{masked\_src}}],
\end{equation}
where $I_{\text{ref}}$ represents the reference image and $I_{\text{masked\_src}}$ is the source image with its insertion region masked.
We complement this visual input with a binary mask $M_{\text{diptych}}$ that designates the reference image region (left panel) with zeros and the insertion region (right panel) with ones:
\begin{equation}
    M_{\text{diptych}} = [\mathbf{0}_{h \times w};  M],
\end{equation}
where $\mathbf{0}_{h \times w}$ has the same dimensions as each panel and M represents the insertion region. This structure provides clear spatial guidance while maintaining contextual relationships between the reference and target regions.

\noindent \textbf{\texttt{Text-Prompt Triptych}.}
For text-prompted editing, we employ a three-panel structure (triptych) consisting of the processed reference image, the unmodified source image, and a fully masked region to be filled:
\begin{equation}
    I_{\text{triptych}} = [R_{\text{seg}}(I_{\text{ref}}); I_{\text{src}};~\emptyset~],
\end{equation}
where $I_{\text{src}}$ represents the source image and $~\emptyset~$ is the empty region to be generated.
Similarly, we create a corresponding binary mask $M_{\text{triptych}}$ that marks the reference and source regions with zeros and the generation region with ones:
\begin{equation}
    M_{\text{triptych}} = [\mathbf{0}_{h \times w}; \mathbf{0}_{h \times w}; \mathbf{1}_{h \times w}],
\end{equation}
where each component has the same dimensions as its corresponding panel.

\begin{table*}[t]
    \small 
    \setlength\tabcolsep{11pt}
    \setlength{\aboverulesep}{0pt}
    \setlength{\belowrulesep}{0pt}
    \centering
    \captionsetup{skip=1pt}
    \caption{\textbf{Quantitative comparison on mask-prompt object insertion tasks.} Evaluations conducted on AnyInsertion and DreamBooth datasets show that our Insert Anything model significantly outperforms existing methods (AnyDoor, MimicBrush, ACE++). The best and second-best results are demonstrated in \textbf{bold} and \underline{underlined}, respectively. ↑/↓ indicate higher/lower is better. Insert Anything achieves state-of-the-art performance across all metrics.}
    \label{tab:mask_comparison_object}
    \begin{tabular}{l|cccc|cccc}
        \toprule
        \multirow{2}{*}{Methods} & \multicolumn{4}{c|}{AnyInsertion (Object)} & \multicolumn{4}{c}{DreamBooth}\\
        \cline{2-9}
        & PSNR $\uparrow$ & SSIM $\uparrow$ & LPIPS $\downarrow$ & FID $\downarrow$ & PSNR $\uparrow$ & SSIM $\uparrow$ & LPIPS $\downarrow$ & FID $\downarrow$\\
        \midrule
        AnyDoor~\cite{chen2024anydoor}& \underline{21.39} & \underline{0.7648} & 0.1831 & 67.99 & 16.68 & 0.5898 & 0.3029 & 95.14 \\
        MimicBrush~\cite{chen2024zero}& 20.80 & 0.7371 & 0.2178 & 67.19 & \underline{18.20} & \underline{0.6039} & 0.2849 & 88.59 \\
        ACE++~\cite{mao2025ace++}& 18.96 & 0.6922 & \underline{0.1485} & \underline{40.11} & 18.06 & 0.5695 & \underline{0.1823} & \underline{64.39} \\
        \rowcolor{LightBlue} Ours & \textbf{26.40} & \textbf{0.8791} & \textbf{0.0820} & \textbf{28.31} & \textbf{21.95} & \textbf{0.7820} & \textbf{0.1350} & \textbf{47.09}  \\
        \bottomrule
    \end{tabular}
    \vspace{-10pt}
\end{table*}

\begin{table}[t]
    \small
    \setlength\tabcolsep{6pt}
    \setlength{\aboverulesep}{0pt}
    \setlength{\belowrulesep}{0pt}
    \centering
    \captionsetup{skip=1pt}
    \caption{\textbf{Quantitative comparison on text-prompt object insertion.} Evaluation on AnyInsertion dataset demonstrates that our Insert Anything model substantially surpasses the baseline method (AnyEdit) across all key metrics.}
    \label{tab:text_comparison_object}
    \begin{tabular}{l|cccc}
        \toprule
        \multirow{2}{*}{Methods} & \multicolumn{4}{c}{AnyInsertion (Object)}\\
        \cline{2-5}
        & PSNR $\uparrow$ & SSIM $\uparrow$ & LPIPS $\downarrow$ & FID $\downarrow$ \\
        \midrule
        AnyEdit~\cite{yu2024anyedit}& 12.74 & 0.5488  &  0.3473 & 226.25 \\
        \rowcolor{LightBlue} Ours & \textbf{17.17} & \textbf{0.6678} & \textbf{0.2011} & \textbf{95.90}\\
        \bottomrule
    \end{tabular}
    \vspace{-10pt}
\end{table}

\subsection{Multiple Control Modes}
\label{sec:method_dit_inpainting}
Our framework supports two control modes for image insertion: mask-prompt and text-prompt. These modes enable flexible, task-specific editing by allowing users to specify insertion regions either manually through masks or via textual descriptions. To seamlessly integrate these input modalities, we leverage the multimodal attention mechanism of DiT~\cite{peebles2023scalable}, utilizing two dedicated branches: an image branch and a text branch.

In our framework, the image branch handles visual inputs, including the reference image, source image, and corresponding masks. These inputs are encoded into feature representations and concatenated with noise along the channel dimension to prepare for generation. In parallel, the text branch encodes the textual description to extract semantic guidance for image editing. The outputs from both branches are then fused via multimodal attention, enabling the model to jointly attend to visual and textual cues. This integration is formalized as:

\begin{equation}    
    Q = [Q_{t}; Q_{i}], K = [K_{t}; K_{i}], V = [V_{t}; V_{i}],
    \label{eq:attn_qkv}
\end{equation}
\begin{equation}
    \text{MMA}(Q, K, V) = \text{softmax}\left(\frac{QK^T}{\sqrt{d}}\right)V,
    \label{eq:attn_eq}
\end{equation}
where $[;]$ represents the concatenation operation, and $Q$, $K$, and $V$ are the query, key, and value components of the attention mechanism. The following describes how the attention mechanism operates under the two control modes.

\noindent \textbf{Mask-Prompt.}
In mask-prompted editing, the insertion region in the source image is specified using a binary mask, as described in \S\ref{sec:method_in_context}. This mask, along with the VAE-processed diptych, is concatenated with noise along the channel dimension and fed into the image branch of the DiT model. Simultaneously, semantic features from the reference image are extracted using a CLIP image encoder and passed into the text branch to provide contextual guidance.

\noindent \textbf{Text-Prompt.}
In text-prompted editing, the insertion is guided by a textual description. The reference image informs the desired modifications, while the text prompt specifies the changes. The source image is modified accordingly to reflect the changes described in the text.
To implement this, we design a specialized prompt template:
\textit{"A triptych with three side-by-side images. On the left is a photo of \texttt{[label]}; on the right, the scene is exactly the same as in the middle but \texttt{[instruction]} on the left."} This structured prompt provides semantic context, where the \texttt{[label]} identifies the reference element type, and the \texttt{[instruction]} specifies the modification. The input is processed through the text encoder, which provides guidance to the text branch of DiT.
The triptych structure (described in \S\ref{sec:method_in_context}) is processed by VAE and fed into the image branch, and the text token is concatenated with the image features to enable joint attention between the branches.
\section{Experiments}
\subsection{Experimental Setup}

\noindent \textbf{Implementation Details.}
Our method builds upon FLUX.1 Fill [dev], an inpainting model based on DiT architecture. The framework integrates a T5~\cite{raffel2020exploring} text encoder and a SigLIP~\cite{zhai2023sigmoid} image encoder, fine-tuned using LoRA with a rank of 256.
For training, we maintained a batch size of 8 for mask prompts and 6 for text prompts, with all images processed at a resolution of 768×768 pixels. We employed the Prodigy optimizer~\cite{mishchenko2023prodigy} with safeguard warmup and bias correction enabled, applying a weight decay of 0.01. All experiments were conducted on a cluster of 4 NVIDIA A800 GPUs (80GB each).
Our AnyInsertion dataset served as our primary training set. We trained the model for 5000 steps across both prompt types (mask and text). For the sampling process, we performed denoising over 50 iterations. Our training loss function follows the flow matching~\cite{lipman2022flow}.

\begin{table*}[t]
    \small 
    \setlength\tabcolsep{11pt}
    \setlength{\aboverulesep}{0pt}
    \setlength{\belowrulesep}{0pt}
    \centering
    \captionsetup{skip=1pt}
    \caption{\textbf{Quantitative comparison on mask-prompt garment insertion.} Evaluation results on AnyInsertion and VTON-HD datasets indicate that our Insert Anything model consistently outperforms existing methods (OOTDiffusion, CatVTON) across all metrics.}
    \label{tab:mask_comparison_garment}
    \small
    \resizebox{\textwidth}{!}{
    \begin{tabular}{l|cccc|cccc}
        \toprule
        \multirow{2}{*}{Methods} & \multicolumn{4}{c|}{AnyInsertion (Garment)} & \multicolumn{4}{c}{VTON-HD}\\
        \cline{2-9}
        & PSNR $\uparrow$ & SSIM $\uparrow$ & LPIPS $\downarrow$ & FID $\downarrow$ & PSNR $\uparrow$ & SSIM $\uparrow$ & LPIPS $\downarrow$ & FID $\downarrow$\\
        \midrule
        ACE++~\cite{mao2025ace++}& 18.11 & 0.7507 & 0.1086 & \underline{35.62} & 17.48 & 0.7634 & 0.1107 & 28.96 \\
        Ootdiffusion~\cite{xu2024ootdiffusion}& 18.07 & 0.8151 & 0.0970 & 87.38 & 21.63 & 0.8643 & 0.0605 & 28.36 \\
        CatVTON~\cite{chong2024catvton}& \underline{23.50} & \underline{0.8477} & \underline{0.0607} & 36.62 & \underline{25.64} & \underline{0.8903} & \underline{0.0513} & \underline{24.80} \\
        \rowcolor{LightBlue} Ours & \textbf{23.78} & \textbf{0.8665} & \textbf{0.0522} & \textbf{28.54} & \textbf{26.10} & \textbf{0.9161} & \textbf{0.0484} & \textbf{19.51}  \\
        \bottomrule
    \end{tabular}}
    \vspace{-10pt}
\end{table*}

\begin{figure*}[t]
  \centering
  \captionsetup{skip=1pt}
  \includegraphics[width=0.90\textwidth]{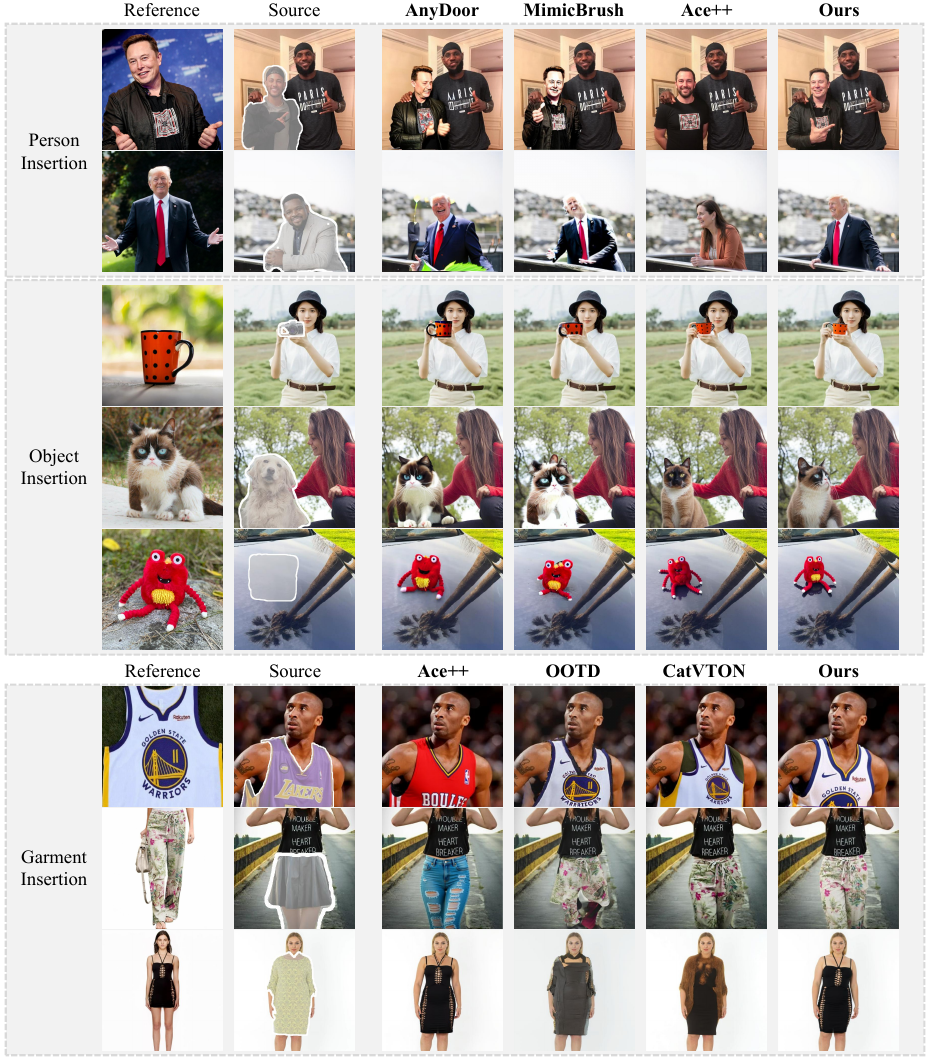}
\caption{\textbf{Qualitative comparison of mask-prompt image insertion results.} Our method consistently preserves identity and maintains visual coherence across diverse insertion tasks (person, object, and garment) compared to existing methods (AnyDoor~\cite{chen2024anydoor}, MimicBrush~\cite{chen2024zero}, Ace++~\cite{mao2025ace++}, OOTD~\cite{xu2024ootdiffusion}, CatVTON~\cite{chong2024catvton}).}
   \label{fig:qualitative_res_mask_prompt}
   \vspace{-12pt}
\end{figure*}

\noindent \textbf{Test Datasets.}
We evaluate our method on three diverse datasets: Insert Anything, DreamBooth~\cite{ruiz2023dreambooth}, and VTON-HD~\cite{choi2021viton}. From our proposed Insert Anything dataset (described in \S\ref{sec:dataset_info}), we select 40 samples for object insertion evaluation, 30 samples for garment insertion, and 30 samples for person insertion (simple scene insertions). For DreamBooth evaluation, we construct a test set comprising 30 group images, where one image serves as the reference and another as the target image. Additionally, we incorporate VTON-HD, which is the standard benchmark for virtual try-on applications and garment insertion tasks.

\noindent \textbf{Metrics.} In experiments, evaluation metrics include Peak Signal-to-Noise Ratio (PSNR), Structural Similarity Index (SSIM), Learned Perceptual Image Patch Similarity (LPIPS), and Fréchet Inception Distance (FID)~\cite{heusel2017gans}.

\noindent \textbf{Baselines.}
We compare against state-of-the-art methods across insertion tasks. For object and person insertion with mask prompts, we benchmark against AnyDoor~\cite{chen2024anydoor}, MimicBrush~\cite{chen2024zero}, and ACE++\cite{mao2025ace++}. For text-prompt object insertion, we compare with AnyEdit\cite{yu2024anyedit}, the current leading open-source method supporting this functionality. For garment insertion, we evaluate against both unified frameworks (ACE++\cite{mao2025ace++}) and specialized methods (OOTDiffusion\cite{xu2024ootdiffusion} and CatVTON~\cite{chong2024catvton}), enabling comprehensive comparison against both task-agnostic and task-specific approaches.

\subsection{Quantitative Results}

\noindent \textbf{Object Insertion Results.} As shown in Tables~\ref{tab:mask_comparison_object} and \ref{tab:text_comparison_object}, Insert Anything consistently outperforms existing methods across all metrics for both mask-prompt and text-prompt object insertion. For mask-prompt insertion, our approach substantially improves SSIM from 0.7648 to 0.8791 on AnyInsertion and from 0.6039 to 0.7820 on DreamBooth. For text-prompt insertion, we achieve a reduction in LPIPS from 0.3473 to 0.2011, indicating significantly better perceptual quality. These improvements demonstrate our model's superior ability to preserve object identity while maintaining integration with the target context.

\noindent \textbf{Garment Insertion Results.} Table~\ref{tab:mask_comparison_garment} shows Insert Anything's consistent superiority over both unified frameworks and specialized garment insertion methods across all metrics on both evaluation datasets. On the widely used VTON-HD benchmark, we improve upon the previous best results from specialized methods, reducing LPIPS from 0.0513 to 0.0484, while simultaneously achieving substantial improvements in PSNR (26.10 \vs 25.64) and SSIM (0.9161 \vs 0.8903). The performance gap widens further when compared to unified frameworks like ACE++, highlighting our approach's effectiveness in combining task-specific quality with a unified architecture.

\begin{table}[t]
    \small
    \setlength\tabcolsep{6pt}
    \setlength{\aboverulesep}{0pt}
    \setlength{\belowrulesep}{0pt}
    \centering
    \captionsetup{skip=1pt}
    \caption{\textbf{Quantitative comparison with state-of-the-art methods on person insertion.} Evaluations on AnyInsertion dataset demonstrate that our Insert Anything Model consistently achieves superior results across all metrics.}
    \label{tab:mask_comparison_person}
    \begin{tabular}{l|cccc}
        \toprule
        \multirow{2}{*}{Methods} & \multicolumn{4}{c}{AnyInsertion (Person)}\\
        \cline{2-5}
        & PSNR $\uparrow$ & SSIM $\uparrow$ & LPIPS $\downarrow$ & FID $\downarrow$ \\
        \midrule
        AnyDoor~\cite{chen2024anydoor}& 14.71 & 0.6807 & 0.3613 & 217.17\\
        MimicBrush~\cite{chen2024zero}& \underline{20.58} & \underline{0.7654} & 0.2125 & 108.26\\
        ACE++~\cite{mao2025ace++}& 19.21 & 0.7513 & \underline{0.1529} & \underline{66.84}\\
        \rowcolor{LightBlue} Ours & \textbf{23.85} & \textbf{0.8457} & \textbf{0.1269} & \textbf{52.77}\\
        \bottomrule
    \end{tabular}
    \vspace{-10pt}
\end{table}

\noindent \textbf{Person Insertion.}
Table~\ref{tab:mask_comparison_person} shows that Insert Anything significantly outperforms existing methods across all metrics for person insertion on AnyInsertion dataset. Our approach achieves notable improvements in structural similarity (SSIM: 0.8457 \vs 0.7654) and perceptual quality (FID: 52.77 \vs 66.84) compared to the previous best results. These improvements are particularly significant considering the challenge of preserving human identity during the insertion process.

\vspace{-5pt}
\begin{figure}[t]
  \centering
  \captionsetup{skip=1pt}
  \includegraphics[width=\linewidth]{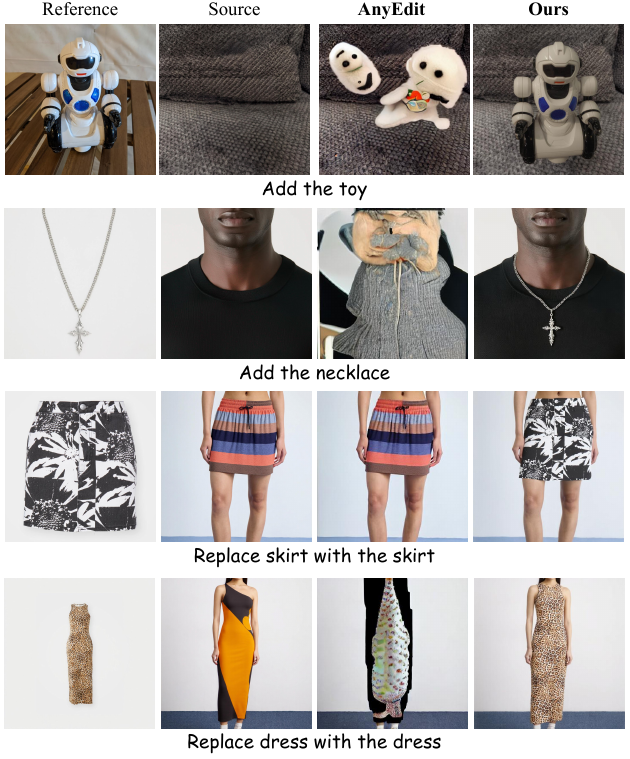}
   \caption{\textbf{Qualitative comparisons with AnyEdit~\cite{yu2024anyedit} on text-prompt object and garment insertion.} See Table~\ref{tab:text_comparison_object} for quantitative results.}
   \label{fig:qualitative_res_object_text}
   \vspace{-12pt}
\end{figure}

\subsection{Qualitative Results}

Fig.~\ref{fig:qualitative_res_mask_prompt} presents visual comparisons across diverse insertion scenarios for all three task categories. Our qualitative analysis reveals several key advantages of Insert Anything:

\noindent \textbf{Object Insertion Quality.}
Insert Anything consistently demonstrates superior preservation of reference object identity while achieving more natural integration within target contexts. This is particularly evident in complex scenarios involving object-person and object-object interactions, where our approach maintains fine-grained details while ensuring contextual harmony. Similarly, Figure~\ref{fig:qualitative_res_object_text} illustrates our model's superior text-guided insertion capabilities compared to AnyEdit, with noticeably better detail preservation.

\noindent \textbf{Garment Insertion Quality.}
Our method excels in challenging garment insertion scenarios, including preserving logos and text on clothing items and handling significant shape transformations (\eg, switching a skirt for pants). Insert Anything demonstrates superior detail preservation and more natural fit compared to both task-specific methods and unified frameworks, highlighting the effectiveness of our approach in balancing specialization with generalization.

\noindent \textbf{Person Insertion Quality.}
Insert Anything exhibits superior performance in complex person insertion scenarios, including person-person interactions, person-animal interactions, and person-object interactions. Our model consistently preserves individual identities while achieving more natural integration with surrounding contexts.

\subsection{Ablation Study}
We conducted an ablation study on mask prompt image insertion. The values in Table~\ref{tab:ablation_res} represent the weighted average results for person insertion, object insertion, and garment insertion, with weights based on the proportions of each category in the mask-prompt test set. Specifically, the test set consists of 40 pairs for object insertion, 30 pairs for garment insertion, and 30 pairs for person insertion. Therefore, the weighted values are object insertion:garment insertion:person insertion = 4:3:3. More ablation results are provided in the supplementary materials.

\begin{figure}[t]
  \centering
  \captionsetup{skip=1pt}
  \includegraphics[width=\linewidth]{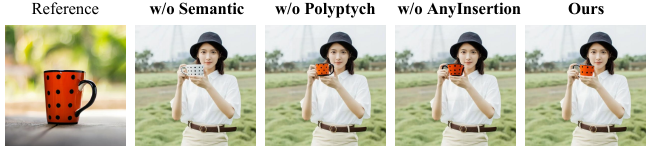}
   \caption{\textbf{Ablation study on mask prompt insertion}. Omitting polyptych in-context, semantic guidance, or AnyInsertion data degrades visual fidelity, sacrificing fine details and semantic cues.}
   \label{fig:qualitative_ablation}
   \vspace{-5pt}
\end{figure}

\begin{table}[t]
\small
\setlength\tabcolsep{4pt}
\centering
\captionsetup{skip=1pt}
\caption{\textbf{Ablation results on mask-prompt insertion.}}
\label{tab:ablation_res}
\begin{tabular}{lcccc}
    \toprule
    Methods & PSNR $\uparrow$ & SSIM $\uparrow$ & LPIPS $\downarrow$ & FID $\downarrow$ \\
    \midrule
    w/o In-Context Edting& 23.37 & 0.8495 & 0.1095 & 59.97 \\
    w/o Semantic Guidance& 24.57 & 0.8635 & 0.0876 & 39.63 \\
    w/o AnyInsertion & 23.60 & 0.8483 & 0.1129 & 61.03 \\
    Ours & \textbf{24.85} & \textbf{0.8653} & \textbf{0.0865} & \textbf{35.72} \\
    \bottomrule
\end{tabular}
\vspace{-10pt}
\end{table}

\noindent \textbf{In-Context Editing.} As shown in Fig.~\ref{fig:qualitative_ablation}, when we remove the polyptych in-context editing during training, the generated images fail to retain fine-grained details (\eg, textures) from the reference image. This results in a significant drop in PSNR, SSIM, and LPIPS values in Table~\ref{tab:ablation_res}, demonstrating the effectiveness of polyptych in-context editing in preserving high-frequency details.

\noindent \textbf{Semantic Guidance.} As shown in Fig.~\ref{fig:qualitative_ablation}, when we remove the semantic guidance for the reference image during training, the generated images lose high-level semantic information (\eg, color) from the reference image. This indicates that semantic guidance plays a crucial role in retaining the coarse, high-level semantic features.

\noindent \textbf{AnyInsertion.} When we remove our custom training data and rely solely on a training-free model for inference, Table~\ref{tab:ablation_res} shows a noticeable drop in all evaluation metrics. Moreover, Fig.~\ref{fig:qualitative_ablation} illustrates that the model's ability to preserve facial details in person insertion is compromised.

\section{Conclusion}
This paper introduces Insert Anything, a unified framework for reference-based image insertion that overcomes the limitations of specialized approaches by supporting both mask- and text-guided control across diverse insertion tasks. Leveraging our newly developed AnyInsertion dataset with 120K prompt-image pairs and the capabilities of DiT architecture, we implement an innovative in-context editing mechanism with diptych and triptych prompting strategies that effectively preserve identity features while maintaining visual harmony between inserted elements and target scenes. Extensive experiments on three benchmarks demonstrate that our method consistently outperforms state-of-the-art methods across person, object, and garment insertion, establishing a new standard for reference-based image editing and providing a versatile solution for real-world creative applications.

{
    \small
    \bibliographystyle{ieeenat_fullname}
    \bibliography{main}
}

\clearpage
\setcounter{page}{1}
\maketitlesupplementary
\appendix

This appendix provides supplementary materials to complement the main manuscript. In \S\ref{sec:more_implement_detail}, we detail additional implementation aspects, including the hybrid masking approach and Adaptive Crop Strategy. Further details on the dataset construction process are presented in \S\ref{sec:more_dataset}. \S\ref{sec:more_ablation} offers additional ablation results and analysis. Subsequently, \S\ref{sec:future_directions} outlines potential directions for future work. Lastly, \S\ref{sec:social} discusses the broader societal impacts of Insert Anything.

\section{More Implementation Details}\label{sec:more_implement_detail}

\noindent\textbf{Hybrid Masking Approach.}
The application of masks during training significantly impacts model performance, particularly when handling mask prompts. We identified limitations in existing approaches: training exclusively with segmentation masks restricts the model's ability to handle free-form masks encountered in real-world applications, while training solely with box masks can hinder the model's ability to use masks for pose guidance and reduce adaptability to non-rectangular shapes.

To address these limitations, we developed a hybrid masking strategy tailored to different insertion categories:
For object and garment categories, we employ box masks with Bessel curve-based shape augmentation~\cite{yang2023paint}, introducing greater variation to improve adaptability across different scenarios. For the person category, we utilize a combination of box masks, augmented box masks, and dilated segmentation masks to provide enhanced pose guidance.

This category-specific approach ensures that the model can effectively handle both precise and free-form masks while maintaining the ability to use mask information for appropriate pose and orientation guidance.

\noindent\textbf{Adaptive Crop Strategy.}
During inference, when an inserted element occupies only a small region of the target image, high-frequency details (such as facial features or object textures) are susceptible to loss. To preserve these details, we implement an adaptive crop-and-zoom strategy.

The process begins by localizing the editing region using either the mask or text prompt. We then crop the editing region along with surrounding pixels to create a new target image, effectively increasing the proportion of the editing region and ensuring it receives more attention during processing.

The crop size significantly impacts the outcome: too small a crop may lack sufficient context for coherent integration, while too large a crop may fail to adequately emphasize the editing region. Our adaptive crop strategy addresses this challenge by dynamically adjusting the crop size based on the editing region's proportional area.

Let \(r\) represent the area ratio of the editing region (a value between 0 and 1) and \(T\) a predefined threshold. When \(r \ge T\), we use the entire image (crop factor = 1). When \(r < T\), we crop a region centered on the editing area with proportionally more context for smaller regions. The crop factor function \(f(r)\) is defined as:

\begin{equation}
  f(r) = \begin{cases}
    \beta + \dfrac{1-\beta}{T}\,r, & \text{if } r < T, \\
    1, & \text{if } r \ge T,
    \end{cases}
  \label{eq:adaptive_crop}
\end{equation}

where:
\begin{itemize}
    \item \(r\) is the area ratio of the editing region,
    \item \(T\) is a predefined threshold,
    \item \(\beta\) (with \(0 < \beta < 1\)) is the minimum crop area fraction when \(r = 0\).
\end{itemize}

In our experiments, we set \(\beta = 0.6\) and \(T = 0.1\). This adaptive approach ensures appropriate emphasis on the editing region while maintaining sufficient context for coherent integration, regardless of the region's size.

\section{More Dataset Information}
\label{sec:more_dataset}

\noindent \textbf{Part I: Data from E-commerce Websites.}
We initiated data collection on various e-commerce websites to acquire a rich and diverse set of reference objects along with corresponding images featuring human subjects. During the collection phase, we gathered a wide range of images related to reference objects, covering categories such as apparel and accessories. To accurately filter for images containing people, we employed human detection techniques (\eg, insightface~\cite{guo2021sample}) to identify images with human presence and exclude those without, completing the first round of screening.

Since a single e-commerce webpage often contains multiple images of the same reference object from different viewpoints, we introduced the LightGlue~\cite{lindenberger2023lightglue} model for similarity detection. In practice, the curated set of human-containing images is used as input, and for each image, LightGlue selects the reference image most similar to the source image among the various viewpoints.

\noindent \textbf{Part II: Data from the HumanVid Dataset.}
The second portion of our data is sourced from the HumanVid~\cite{wang2024humanvid} video dataset. We first apply the Laplacian algorithm to filter out frames with motion blur by computing the Laplacian variance of the first 10 grayscale frames, using the maximum value as a threshold—since initial frames typically exhibit minimal motion blur. Additionally, we select paired data where the person’s overall pose changes while the head pose remains stable. Specifically, the first clear frame is designated as the starting keyframe; then, using a head pose model, we detect the head pose angles in subsequent frames. The ending keyframe is chosen as the frame that is as far apart as possible from the starting keyframe while keeping the head pose within a reasonable range. This process effectively mitigates motion blur and ensures the acquisition of paired data with varying body poses but stable head orientations.

\noindent \textbf{Part III: Data from MVImgNet.}
The third portion of our data is drawn from the MVImgNet~\cite{yu2023mvimgnet} dataset, which comprises multi-view images of objects. To ensure that the selected data exhibits moderate viewpoint variation without excessive differences that may hinder learning, we extract data at appropriately spaced intervals. This strategy captures object features across different viewpoints while keeping the variations within a manageable and controllable range.

\vspace{-5pt}
\begin{figure}[h!]
  \centering
  \captionsetup{skip=1pt}
  \includegraphics[width=\linewidth]{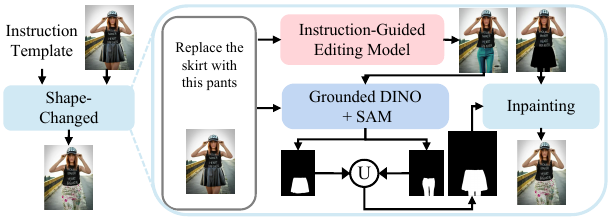}
   \caption{\textbf{Shape Changing Data Construction.} Image pairs are processed by instruction-guided editing, segmentation (Grounded DINO~\cite{liu2024grounding}+SAM~\cite{kirillov2023segment}), and inpainting refinement.} 
   \label{fig:data-shape-change-construction}
   \vspace{-8pt}
\end{figure}

\noindent \textbf{Handling Shape Changes.}
As depicted in Figure~\ref{fig:data-shape-change-construction}, our text-to-mask module effectively addresses shape changes by initially applying a text-based editing model. The module then extracts and combines the masks from both the edited and unedited images. This combined mask serves as a guide for the inpainting model, enabling it to generate source images that ensure consistency in the regions that are not modified, while preserving the integrity of the original content.

\section{More Ablation Results}\label{sec:more_ablation}

\begin{table}[htbp]
\small
\setlength\tabcolsep{4pt}
\centering
\captionsetup{skip=1pt}
\caption{Ablation results on person insertion in mask prompt editing.}
\label{tab:more_ablation_res}
\begin{tabular}{lcccc}
    \toprule
    Methods & PSNR $\uparrow$ & SSIM $\uparrow$ & LPIPS $\downarrow$ & FID $\downarrow$ \\
    \midrule
    w/o Semantic Guidance& 24.57 & 0.8635 & 0.0876 & 39.63 \\
    w/o Polyptych In-Context& 23.37 & 0.8495 & 0.1095 & 59.97 \\
    w/o IA Data & 23.60 & 0.8483 & 0.1129 & 61.03 \\
    w/o Adaptive Crop & 23.89 & 0.8459 & 0.0977 & 37.75 \\
    w/ Instance Mask & 24.46 & 0.8648 & 0.0870 & 37.15 \\
    w/ Box Mask & \textbf{24.88} & 0.8637 & 0.0905 & 39.51 \\
    Ours & 24.85 & \textbf{0.8653} & \textbf{0.0865} & \textbf{35.72} \\
    \bottomrule
\end{tabular}
\end{table}

\vspace{-5pt}
\begin{figure}[h!]
  \centering
  \captionsetup{skip=1pt}
  \includegraphics[width=\linewidth]{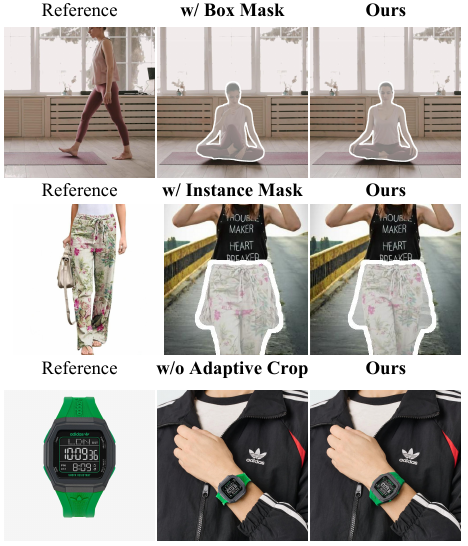}
   \caption{\textbf{Ablation study on adaptive cropping and hybrid masking strategies.} Removing adaptive cropping reduces detail preservation for small objects, while using only box or instance masks compromises either pose adaptability or mask-guided precision. Our combined strategy achieves balanced and high-quality insertion results.}
   \label{fig:more_abla}
   \vspace{-8pt}
\end{figure}

\noindent \textbf{Adaptive Crop Strategy.} To validate the effectiveness of our adaptive crop strategy, we removed dynamic adaptive cropping during inference. As shown in Fig.~\ref{fig:more_abla}, the model’s ability to handle small objects and preserve high-frequency details degrades, confirming the benefit of adaptive cropping.

\noindent \textbf{Hybrid Masking.} To assess the necessity of our hybrid masking approach, we modified the masks used during training by testing two baselines: one using only instance masks and another using only box masks. As shown in Fig.~\ref{fig:more_abla}, training exclusively with instance masks leads the model to completely fill the mask region, preventing flexible, content-adaptive adjustments. This indicates that box masks are crucial for promoting adaptability during training. Conversely, when training solely with box masks, the model’s ability to guide pose based on the mask diminishes; even with the provided mask, the final output fails to conform to the desired pose. These observations suggest that instance masks are beneficial for ensuring the output follows the mask’s pose guidance.

\section{Future Directions}
\label{sec:future_directions}
While Insert Anything demonstrates substantial improvements over existing approaches, our research has identified several promising directions for future development. First, incorporating physics-conformant constraints into the editing process would significantly enhance the realism of generated images. Such constraints would ensure that edited elements respect physical laws, producing more naturalistic results when placing objects or adjusting scene compositions.

Second, the implementation of a Mixture of Experts (MoE) architecture presents an opportunity to enhance model performance through specialization. By developing expert models for different editing scenarios, we can improve both efficiency and accuracy across diverse tasks. This approach would enable more nuanced handling of specific editing challenges, from person insertion to object manipulation, while maintaining computational efficiency.

Third, expanding the conditioning capabilities of our polyptych format offers potential for more sophisticated editing control. Additional conditioning types could enable finer-grained control over editing operations, allowing for more precise and context-aware modifications. This enhancement would particularly benefit complex editing scenarios requiring detailed attention to spatial relationships and semantic consistency.

These future directions align with our goal of developing increasingly sophisticated and user-friendly image editing tools. By addressing these areas, we can further advance the capabilities of reference-based image editing while maintaining the computational efficiency and accessibility that characterize Insert Anything.

\section{Broader Impacts}\label{sec:social}

\subsection{Positive Societal Impacts}
\noindent \textbf{Enhanced Creative Expression}
Insert Anything empowers content creators, digital artists, and designers to produce high-quality, realistic edits, thereby fostering innovation and artistic expression.

\noindent \textbf{Increased Accessibility}
By providing a unified, controllable image insertion framework that simplifies complex editing tasks, non-expert users can more easily generate professional-level content for social media, advertising, and e-commerce.

\noindent \textbf{Improved Virtual Try-On and Personalization}
The technology can enhance virtual try-on applications and personalized digital experiences, benefiting online retail and fashion industries by allowing more accurate and appealing visualizations.

\noindent \textbf{Efficiency in Digital Media Production} The unified model reduces the need for multiple specialized tools, streamlining workflows in film production, graphic design, and digital marketing, and potentially lowering costs.

\subsection{Negative Societal Impacts}
\noindent \textbf{Risk of Misuse and Deepfakes} The capability to seamlessly insert elements from one image into another raises concerns about the creation of misleading or deceptive content, such as deepfakes, which could be used to spread misinformation.

\noindent \textbf{Privacy and Ethical Concerns} The technology might be exploited to manipulate personal images without consent, raising ethical issues related to privacy, identity theft, and reputational harm.

\noindent \textbf{Intellectual Property Issues} The ability to insert and modify content from reference images could lead to unauthorized use or misrepresentation of copyrighted material.

\noindent \textbf{Erosion of Trust in Visual Media} As manipulated images become more realistic, it may become increasingly difficult for viewers to distinguish authentic content from edited or fabricated images, potentially undermining trust in visual media.

\end{document}